\pdfoutput=1  
\documentclass[letterpaper,11pt]{article}

\usepackage[letterpaper,left=1.4in,right=1.4in,top=1.1in,bottom=1.25in,
            headheight=19pt,headsep=12pt]{geometry}
\usepackage[T1]{fontenc}
\usepackage[utf8]{inputenc}
\usepackage{lmodern}    
\usepackage{newtxtext}  
\usepackage{newtxmath}  

\usepackage[hyphens]{url}
\usepackage{graphicx}
\usepackage[round]{natbib}
\usepackage[font=small,labelfont=bf]{caption}
\usepackage{ragged2e}
\usepackage{microtype}
\usepackage{amsmath}
\usepackage{amssymb}
\usepackage{booktabs}
\usepackage{multirow}
\usepackage{tabularx}
\usepackage{array}
\usepackage{pifont}
\usepackage{xcolor}
\usepackage{xspace}
\definecolor{inkblue}{HTML}{2F4B7C}
\usepackage[colorlinks=true,allcolors=inkblue,breaklinks=true]{hyperref}

\renewcommand{\arraystretch}{1.12}   
\makeatletter
\renewcommand\section{\@startsection{section}{1}{\z@}%
  {-3.5ex \@plus -1ex \@minus -.2ex}{1.8ex \@plus .2ex}%
  {\normalfont\large\bfseries}}
\renewcommand\subsection{\@startsection{subsection}{2}{\z@}%
  {-2.8ex \@plus -0.8ex \@minus -.2ex}{1.2ex \@plus .2ex}%
  {\normalfont\normalsize\bfseries}}
\renewcommand\paragraph{\@startsection{paragraph}{4}{\z@}%
  {1.8ex \@plus .4ex \@minus .2ex}{-0.65em}%
  {\normalfont\normalsize\bfseries}}
\makeatother

\makeatletter
\def\ps@preprint{%
  \let\@mkboth\@gobbletwo
  \def\@oddhead{\parbox[b]{\textwidth}{%
      \footnotesize\scshape World Tokens\hfill A Preprint\\[1.5pt]
      \rule{\textwidth}{0.3pt}}}%
  \let\@evenhead\@oddhead
  \def\@oddfoot{\hfil\small\thepage\hfil}%
  \let\@evenfoot\@oddfoot
}
\makeatother
\definecolor{markoff}{HTML}{9A9A9A}
\definecolor{okgreen}{HTML}{1FA84C}
\definecolor{failred}{HTML}{E03127}
\newcommand{\cmark}{\ding{51}}
\newcommand{\xmark}{\textcolor{markoff}{\ding{55}}}
\newcommand{\okmark}{\textcolor{okgreen}{\ding{51}}}
\newcommand{\failmark}{\textcolor{failred}{\ding{55}}}
\newcommand{\method}{World Tokens\xspace}
\newcommand{\wm}{\mathcal{W}}

\newcommand{\loss}{\mathcal{L}}

\begin{document}

\begin{center}
  {\LARGE\bfseries \method: Enhancing Embodied Policies with\\[0.35em]
   Training-Time World Modeling\par}
  \vskip 1.6em
  {\large Qu Tang \quad Benhui Zhuang \quad Bo Yuan\textsuperscript{\textdagger}
   \quad Xue Yu \quad Longteng Guo \quad Junlan Feng\par}
  \vskip 0.8em
  {\normalsize JIUTIAN Research \qquad Zhongguancun Academy\par}
  \vskip 0.45em
  {\footnotesize\textsuperscript{\textdagger}Corresponding author.\par}
\end{center}

\vskip 1.8em
\noindent\rule{\textwidth}{0.8pt}
\vskip 0.8em
\begin{center}{\large\bfseries Abstract}\end{center}
\vskip 0.3em
\begingroup
\small
\leftskip=0.045\textwidth \rightskip=0.045\textwidth
\noindent Vision-language-action (VLA) models are a widely adopted paradigm for embodied policies. They excel at efficient closed-loop control but do not explicitly model how physical scenes evolve as a task unfolds. Recently emerging world-action models (WAMs) leverage pretrained video world models to capture spatiotemporal evolution, yet retaining future generation or a large video backbone in the control loop substantially increases inference cost. We introduce World Tokens, an embodied policy architecture built around a World Adapter that bridges visual-language understanding, world-dynamics modeling, and action generation. It uses world modeling during training to enhance the action policy while preserving efficient deployment. Specifically, the World Adapter transforms VLM features into a fixed set of world tokens, which condition a jointly fine-tuned future-video denoiser and simultaneously serve as the action expert's sole visual-language context. This shared conditioning allows gradients from future-video denoising to directly shape the representation used for action prediction, while exclusive routing prevents the policy from bypassing that representation. At deployment, the world-model branch is removed, leaving only the VLM, World Adapter, and action expert, with no online video-model inference. With a 2B backbone and no embodied action pretraining, World Tokens is highly competitive on LIBERO, attains the best reported averages on SIMPLER, substantially improves real-world R1~Pro success over a matched action-only baseline, and generates each action chunk at VLA-level latency.
\par
\endgroup
\vskip 0.9em
\noindent\rule{\textwidth}{0.8pt}
\vskip 2.0em
\thispagestyle{plain}

\section{Introduction}

Vision-language-action (VLA) policies pair a pretrained vision-language model (VLM) with an action expert that maps visual observations and language instructions to continuous robot commands \citep{kim2024openvla,black2024pi0,bjorck2025gr00t,kim2025openvlaoft,black2025pi05}. The VLM provides semantic recognition and instruction grounding, while the action expert supports efficient closed-loop control. However, image--text pretraining provides little direct supervision for action-relevant temporal transitions, such as contact, occlusion, and object displacement. Because appropriate action selection often depends on how the scene is likely to evolve rather than on current appearance alone, a VLA must learn these transitions primarily from robot demonstrations, which are typically much more limited than its visual-language pretraining data.

Pretrained video world models offer complementary supervision by predicting how scenes change over time. World-action models (WAMs) incorporate this capability by jointly generating videos and actions, conditioning actions on predicted futures, or extracting features from a video-denoising backbone \citep{bi2025motus,ye2026dreamzero,kim2026cosmospolicy,ma2026dit4dit,li2026lingbot}. These architectures have shown promising long-horizon and out-of-distribution manipulation performance, but iterative video denoising or a retained multi-billion-parameter backbone can substantially increase control latency.

\begin{figure}[htbp]
  \centering
  \begin{minipage}[c]{0.56\textwidth}
    \includegraphics[width=\linewidth]{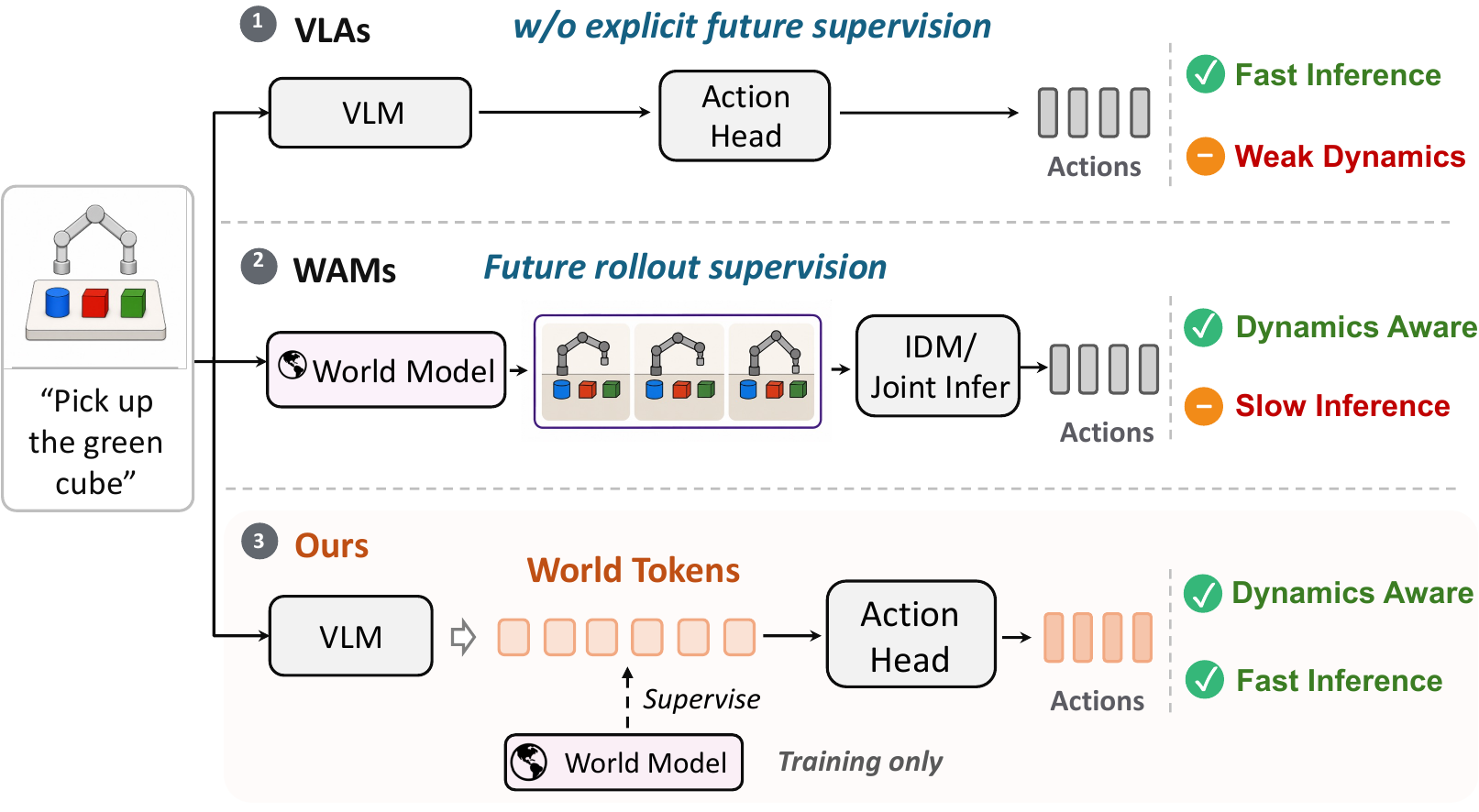}
  \end{minipage}\hfill
  \begin{minipage}[c]{0.40\textwidth}
    \captionsetup{justification=RaggedRight,singlelinecheck=false}%
    \caption{\method combines the fast inference of behavior cloning with the dynamics awareness of world models by moving world modeling to training time. \textbf{(1)} Behavior-cloning VLAs act directly on VLM features: fast, but without dynamics supervision. \textbf{(2)} World-model rollout predicts futures inside the control loop: dynamics-aware, but slow. \textbf{(3)} Our method uses the world model only during training and discards it at deployment.}
  \label{fig:teaser}
  \end{minipage}
\end{figure}

Recent evidence suggests separating the representational benefit of predictive training from the cost of online video inference (Figure~\ref{fig:teaser}). Fast-WAM shows that video co-training can contribute more than explicit future synthesis alone, yet its deployed controller still retains a video DiT together with a large companion action DiT and language encoder \citep{yuan2026fastwam}. In parallel, lighter predictive objectives can improve VLAs without retaining a video backbone at deployment \citep{fang2025motion,sun2026vlajepa}. A key design issue remains: predictive supervision typically acts only through a parallel branch, while the action expert still conditions on the full VLM context. The controller can then simply ignore the subspace shaped by world modeling, and prediction and action can drift apart in representation space. We therefore ask: \emph{how can world modeling better enhance the representation used for embodied control while the world model itself is removed completely after training?}

\begin{figure}[t]
  \centering
  \includegraphics[width=0.94\textwidth]{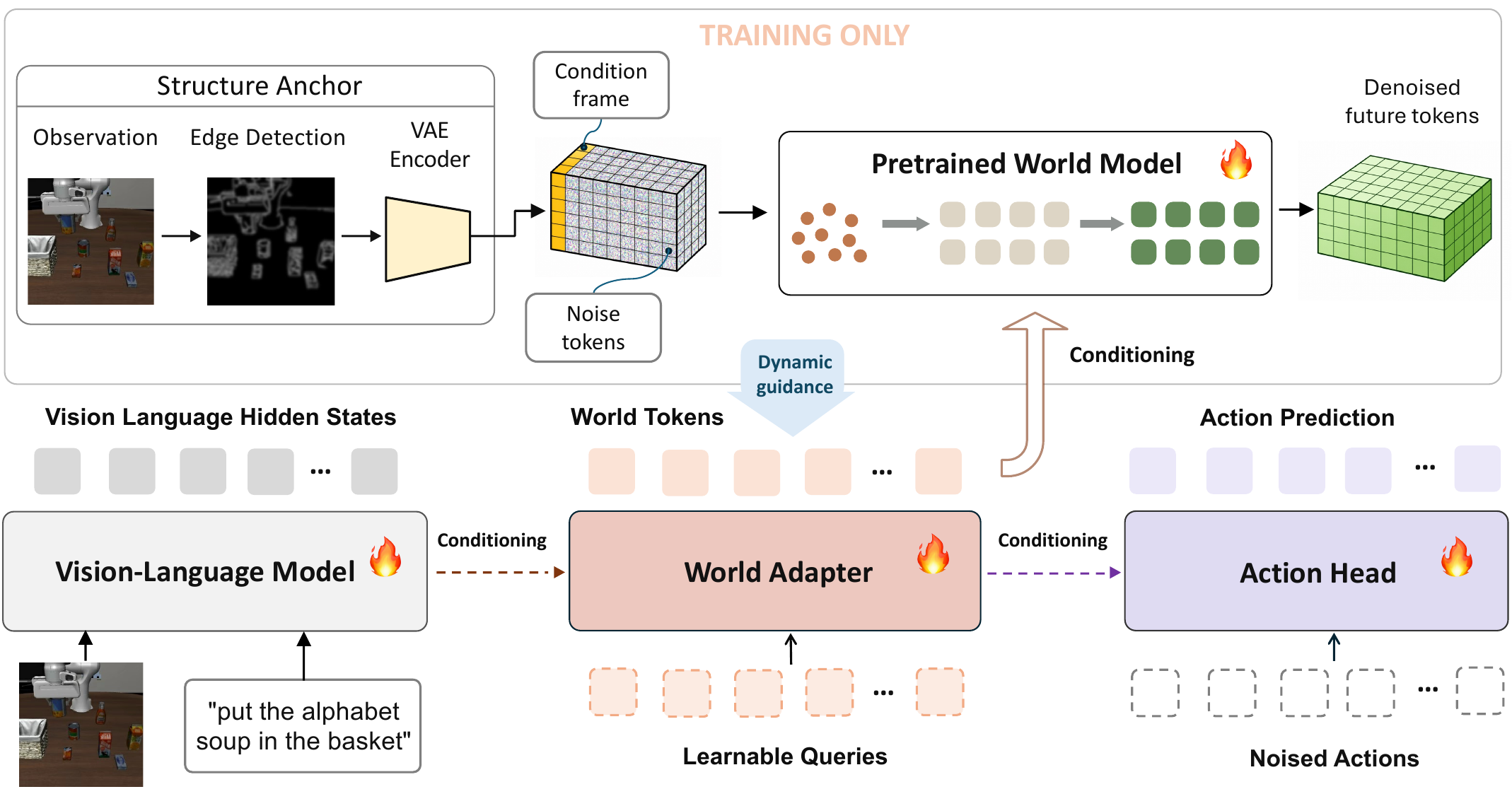}
  \caption{\textbf{Training architecture.} Given multi-view observations and language, the VLM produces multimodal features that the World Adapter maps into a fixed set of world tokens $q_t$. These tokens condition a jointly fine-tuned video denoiser (top) and a flow-matching action expert (bottom right). A Canny edge map of the primary view, encoded as a first-frame latent, supplies appearance-suppressed spatial structure to the video branch. At deployment, the video tokenizer, denoiser, structural anchor, and future targets are removed; the VLM, World Adapter, and action expert remain.}
  \label{fig:method}
\end{figure}

We address this question with a World Adapter (Figure~\ref{fig:method}), an intermediate module that bridges task understanding, dynamics evolution, and action generation (i.e., the VLM, a video world model, and an action DiT). During training, VLM features are expressed through the World Adapter as a fixed set of world tokens, which jointly condition a fine-tuned future-video denoiser and a flow-matching action expert. The action expert has no parallel access to the full VLM sequence; all visual and linguistic context must pass through these tokens. This exclusive routing precludes a direct controller bypass around the future-supervised representation. In parallel, we weaken the world model's first-frame condition to an edge map that retains only scene-layout semantics. This prevents the video branch from relying on full appearance cues, so the world tokens must carry the rich visual-semantic content needed to guide the world model's inference of scene evolution.

Consequently, the video world model is needed only to supervise the world tokens during training and need not enter the closed control loop. At deployment, the full video branch is removed and inference runs entirely through the VLM, World Adapter, and action expert: the control representation has already been shaped by world modeling, while inference latency remains close to that of a VLA. Empirically, \method ranks among the strongest entries on LIBERO, where training and evaluation remain in simulation, and attains the best reported averages on SIMPLER, where policies trained on real data are evaluated in simulation; on a physical R1~Pro robot, it outperforms a matched action-only baseline at comparable latency. Overall, the \method architecture folds world modeling into a deployable control interface, combining dynamics supervision with efficient closed-loop control.

Our contributions are summarized as follows:
\begin{itemize}
    \item We introduce \textbf{\method}, a training-time world-modeling architecture that jointly trains a video world model and a VLA policy. At deployment, the entire video branch is removed, retaining the benefit of predictive supervision at VLA-level inference latency.
    \item We introduce the \textbf{World Adapter}, which maps the VLM sequence into $256$ world tokens that condition the video world model and constitute the action expert's sole visual-language context, so that future-video supervision directly shapes the representation used for control.
    \item Our extensive experiments show that a 2B model without embodied action pretraining reaches 98.2\% on LIBERO and the best performance on SIMPLER-WidowX (71.5\%) and SIMPLER-GoogleRobot (82.1\%), and improves real-robot success from 59.4\% to 76.0\% over a matched baseline.
\end{itemize}

\section{Related Work}

\paragraph{Vision-language-action policies.}
Large-scale robot transformers established the benefits of scaling policy capacity and data diversity across tasks and embodiments \citep{brohan2023rt1,ghosh2024octo}, while RT-2 transferred web-scale visual-language knowledge into robotic control \citep{brohan2023rt2}. Complementary action-generation approaches model multimodal behavior through latent action tokens or diffusion \citep{lee2024vqbet,chi2023diffusionpolicy,liu2025rdt}. Modern VLA models adapt multimodal foundation models to robot control through autoregressive action decoding \citep{kim2024openvla,kim2025openvlaoft} or flow-matching and diffusion action experts conditioned on VLM features \citep{black2024pi0,bjorck2025gr00t,black2025pi05}. We retain this dual-system organization but, rather than conditioning the action expert on the full VLM sequence, route control through a fixed-size token interface jointly constrained by action prediction and future-video denoising.

\paragraph{WAMs.}
Earlier work connected video modeling to control by jointly predicting future images and actions, extracting predictive video representations, learning task-centric latent actions from heterogeneous videos, or generating synthetic robot trajectories \citep{wu2024gr1,hu2025vpp,bu2025univla,jang2025dreamgen}. WAMs more tightly couple video prediction and control through a shared generative backbone. Motus and DreamZero jointly model video and action with large video DiTs \citep{bi2025motus,ye2026dreamzero}; other systems adapt video foundation models for control and planning \citep{kim2026cosmospolicy,ma2026dit4dit} or execute causal future rollouts asynchronously \citep{li2026lingbot}. These methods differ in whether they decode future RGB frames, expose denoising features to the action module, or plan over predicted futures, but all retain substantial video-model computation at deployment. Fast-WAM removes explicit future synthesis yet still evaluates a pretrained video DiT as a world encoder \citep{yuan2026fastwam}.

\paragraph{Predictive world modeling for VLA policies.}
Predictive VLA methods vary in both the target they model and the computation they retain at inference. WorldVLA and UVA jointly generate images and actions \citep{cen2025worldvla,li2025uva}, while DreamVLA predicts dynamic, spatial, and semantic world knowledge before action generation \citep{zhang2025dreamvla}. Other methods use lighter training signals, including future-latent alignment \citep{zheng2025flare}, motion-image diffusion \citep{fang2025motion}, and latent-state prediction of future images \citep{sun2026vlajepa}. Related token formulations either roll compact visual states forward in time or encode discrete world changes \citep{tang2026onewm,zhu2026deltavla}. \method differs in two respects: predictive supervision is sourced from a pretrained video world model rather than a lightweight proxy, and the tokens represent the current context---not a predicted future---and serve as the sole conditioning input to the policy.

\section{Method}

\subsection{Problem Formulation}

We consider vision-language-conditioned robotic manipulation. At control step $t$, the robot observes RGB images from $V$ views, $o_t=(o_t^1,\ldots,o_t^V)$, and receives a language instruction $\ell$. The policy predicts an $H$-step action chunk $a_t^H\equiv(a_t,\ldots,a_{t+H-1})$. During training, each demonstration additionally provides a $T$-frame future RGB clip from the primary view,
$v_t\equiv(o_{t+1}^1,\ldots,o_{t+T}^1)$. The clip is used only as a prediction target: it is never provided to the action policy and is unavailable at evaluation. Our goal is to use $v_t$ to shape the control representation while preserving the standard deployed mapping from $(o_t,\ell)$ to $a_t^H$.

\subsection{Architecture Overview}

\method uses different computational graphs for training and deployment, as shown in Figure~\ref{fig:method}. The deployed policy, shown along the lower path, contains a VLM, a World Adapter, and an action expert. The upper path adds a video world model only for training. A fixed set of world tokens connects the two paths: the action expert uses them to generate actions, while the video model uses a projected version to denoise the demonstrated future clip. The world tokens are the action expert's sole visual-language context, and the video branch receives a Canny edge anchor~\citep{canny1986edge} in place of the full RGB first frame.

\subsection{World Adapter}

\paragraph{Observation and language encoding.}
A pretrained VLM $f_{\phi}$ jointly encodes all image views and the instruction into a variable-length sequence $h_t\in\mathbb{R}^{N_t\times d_h}$ of patch and language tokens. We insert a World Adapter $b_{\psi}$ that maps this sequence into a fixed-size representation:
\begin{equation}
\begin{aligned}
h_t &= f_{\phi}(o_t,\ell),\\
q_t &= b_{\psi}(h_t)\in\mathbb{R}^{K\times d}.
\end{aligned}
\label{eq:world_tokens}
\end{equation}
We term the $K$ output vectors of $q_t$ the world tokens; the name reflects their predictive training objective rather than any predefined slot semantics.

\paragraph{Query-based token extraction.}
The adapter follows the Perceiver resampler design~\citep{jaegle2021perceiver,alayrac2022flamingo}: it maintains $K$ learned queries $Q^{(0)}\in\mathbb{R}^{K\times d}$, and each of its $L$ blocks cross-attends from the current queries to $h_t$, exchanges information among the queries through self-attention, and applies a feed-forward update. Denoting the $j$-th block by $B_{\psi}^{(j)}$, we write
\begin{equation}
\begin{aligned}
Q_t^{(0)} &= Q^{(0)},\\
Q_t^{(j+1)} &= B_{\psi}^{(j)}(Q_t^{(j)},h_t),
\quad j=0,\ldots,L-1,\\
b_{\psi}(h_t) &= Q_t^{(L)}.
\end{aligned}
\label{eq:world_adapter}
\end{equation}
The action and video objectives jointly determine what information the queries retain.

\paragraph{Exclusive action pathway.}
The action expert $g_{\theta}$ is a DiT flow-matching model~\citep{peebles2023dit,lipman2023flow} that generates the action chunk $a_t^H$ conditioned solely on the world tokens:
\begin{equation}
a_t^H \sim g_{\theta}(\cdot \mid q_t).
\label{eq:action_policy}
\end{equation}

\subsection{Training-Time World Modeling}

\paragraph{Video denoiser.}
The training-only video branch is a pretrained video diffusion model, denoted by $\wm_{\omega}$. A lightweight projection $p_{\rho}$ adapts the world tokens to its cross-attention dimension:
\[
q_t^{\mathrm{wm}}=p_{\rho}(q_t).
\]
A frozen VAE encoder $E$ maps the demonstrated future clip to the latent target $z_t=E(v_t)$. During training, $\wm_{\omega}$ denoises noisy versions of $z_t$, using $q_t^{\mathrm{wm}}$ as cross-attention context and the structural anchor defined below as its first-frame condition. Because $q_t^{\mathrm{wm}}$ is derived from $q_t$, the video-denoising gradient directly shapes the same representation consumed by the action expert.

\paragraph{Structural anchor.}
The video branch uses a first-frame reference to anchor the current scene layout. Without this reference, the world tokens must also encode static geometry; with a full RGB frame, the denoiser can rely heavily on appearance persistence, weakening its dependence on the tokens. We therefore replace the RGB reference with a Canny edge map of the primary view, encoded by the same frozen VAE encoder:
\[
c_t=E\big(\mathrm{Canny}(o_t^1)\big).
\]
The resulting latent preserves boundaries and spatial layout while suppressing most color and texture cues. The two conditions are complementary: $c_t$ anchors scene geometry, while $q_t^{\mathrm{wm}}$ supplies task semantics and evolution-relevant content.

\subsection{Joint Training and Deployment}

\paragraph{Training objective.}
Both branches are trained with a conditional denoising objective. For a target $y$, we sample base noise $\epsilon\sim\mathcal{N}(0,I)$ and a noise level $\tau$, and write the noised input as $y_{\tau}$ and the regression target as $u_{\tau}(y,\epsilon)$. For a conditional prediction network $F$, we define
\begin{equation}
\loss_{\mathrm{den}}(F;y\mid c)
=
\mathbb{E}_{\tau,\epsilon}
\left[
\left\|
F(y_{\tau},\tau\mid c)-u_{\tau}(y,\epsilon)
\right\|_2^2
\right].
\label{eq:denoising}
\end{equation}
The two branches differ only in parameterization: the action expert uses flow matching on a linear interpolation path, with $\tau\sim\mathcal{U}[0,1]$, $y_{\tau}=(1-\tau)\epsilon+\tau y$, and $u_{\tau}(y,\epsilon)=y-\epsilon$, while the video branch retains the native diffusion formulation of the underlying video model~\citep{nvidia2025cosmos}. We apply this objective to the demonstrated action chunk and the future-video latent:
\begin{equation}
\begin{aligned}
\loss_{\mathrm{act}}
&=\loss_{\mathrm{den}}\!\left(g_{\theta};a_t^H\mid q_t\right),\\
\loss_{\mathrm{vid}}
&=\loss_{\mathrm{den}}\!\left(\wm_{\omega};z_t
\mid q_t^{\mathrm{wm}},c_t\right),\\
\loss&=\loss_{\mathrm{act}}+\lambda_w\loss_{\mathrm{vid}},
\qquad \lambda_w=0.5.
\end{aligned}
\label{eq:training}
\end{equation}

\paragraph{Gradient paths.}
The action loss updates $f_{\phi}$, $b_{\psi}$, and $g_{\theta}$, forcing the world tokens to retain information needed for control; the video loss updates $f_{\phi}$, $b_{\psi}$, $p_{\rho}$, and $\wm_{\omega}$, encouraging the same tokens to explain the demonstrated scene evolution. Because both gradient paths meet at $q_t$, predictive supervision directly shapes the representation consumed by the policy rather than a detachable auxiliary feature.

\paragraph{Initialization.}
The VLM and the video denoiser are initialized from pretrained checkpoints and fine-tuned jointly. The World Adapter, world-model projection, and action expert are trained from scratch. We fine-tune the video denoiser because robot demonstrations differ from its pretraining data in viewpoint, embodiment, and contact dynamics; a frozen denoiser could provide poorly matched supervision. The tokenizer $E$ remains frozen to keep the target latent space stationary.

\paragraph{Deployment.}
At deployment, the entire video branch---$E$, $p_{\rho}$, and $\wm_{\omega}$---is removed, leaving only $f_{\phi}$, $b_{\psi}$, and $g_{\theta}$; neither the future target $v_t$ nor the structural anchor is required. The policy recomputes $q_t$ at each control step and generates the action chunk with four flow-integration steps. World modeling therefore shapes the control representation during training without adding video-model computation at deployment.

\begin{table}[htbp]
\centering
\small
\setlength{\tabcolsep}{3.1pt}
\resizebox{\textwidth}{!}{%
\begin{tabular}{lcccccccc}
\toprule
\multicolumn{9}{l}{\textbf{Vision-language-action policies (VLAs)}} \\
Method & Backbone Size & Emb. PT & Latency (ms) & Spatial & Object & Goal & Long & Avg. \\
\midrule
OpenVLA \citep{kim2024openvla} & 7B & \cmark & 240 & 84.7 & 88.4 & 79.2 & 53.7 & 76.5 \\
$\pi_{0}$ \citep{black2024pi0} & 3B & \cmark & 73 & 96.8 & 98.8 & 95.8 & 85.2 & 94.1 \\
$\pi_{0.5}$ \citep{black2025pi05} & 2B & \cmark & 56.32 & 98.8 & 98.2 & 98.0 & 92.4 & 96.9 \\
OpenVLA-OFT \citep{kim2025openvlaoft} & 7B & \cmark & 112 & 97.6 & 98.4 & 97.9 & 94.5 & 97.1 \\
VLA-JEPA \citep{sun2026vlajepa} & 2B & \cmark & -- & 96.2 & 99.6 & 97.2 & 95.8 & 97.2 \\
StarVLA \citep{ye2026starvla} & 4B & \xmark & -- & 98.7 & 99.7 & 98.6 & 94.2 & 97.8 \\
World2Act \citep{vuong2026world2act} & 3B & \cmark & -- & 99.5 & \textbf{100.0} & \textbf{98.8} & 94.0 & 98.1 \\
\midrule
\method (ours) & 2B & \xmark & 61.85 & \textbf{99.6} & 98.8 & 97.4 & 97.0 & 98.2 \\
\midrule
\multicolumn{9}{l}{\textbf{WAMs}} \\
Method & Backbone Size & Emb. PT & Latency (ms) & Spatial & Object & Goal & Long & Avg. \\
\midrule
WorldVLA \citep{cen2025worldvla} & 7B & \xmark & -- & 87.6 & 96.2 & 83.4 & 60.0 & 81.8 \\
Fast-WAM \citep{yuan2026fastwam} & 5B & \xmark & 182 & 98.2 & \textbf{100.0} & 97.0 & 95.2 & 97.6 \\
Motus \citep{bi2025motus} & 5B + 2B & \cmark & -- & 96.8 & 99.8 & 96.6 & \textbf{97.6} & 97.7 \\
Cosmos Policy \citep{kim2026cosmospolicy} & 2B & \xmark & 610 & 98.1 & \textbf{100.0} & 98.2 & \textbf{97.6} & 98.5 \\
DiT4DiT \citep{ma2026dit4dit} & 2B & \xmark & 136 & 98.4 & 99.6 & 98.6 & \textbf{97.6} & \textbf{98.6} \\
\bottomrule
\end{tabular}}
\caption{LIBERO success rate (\%) and latency per action chunk (ms). Each suite has 10 tasks; Avg.\ is the unweighted suite mean. Emb.\ PT denotes embodied action pretraining. World2Act discards its world model at deployment and is therefore listed with VLAs. Latency for \method and $\pi_{0.5}$ is measured on our hardware; other values are quoted from source papers for reference, with measurement details in Appendix~\ref{supp:latency}. Bold marks the best success rate in each column.}
\label{tab:libero}
\end{table}

\section{Experiments}

Our evaluation addresses three questions: does training-time world modeling help once the video branch is removed at deployment? Which components drive the improvement? Does the policy remain effective beyond LIBERO? We answer them with LIBERO results, controlled ablations, SIMPLER, and R1 Pro experiments.

\subsection{Experimental Setup}

\paragraph{Implementation.}
All experiments use the same configuration: Qwen3-VL-2B-Instruct~\citep{bai2025qwen3vl} as the VLM, a World Adapter with $K{=}256$ learned queries, a DiT-B action expert, and Cosmos Predict2.5-2B~\citep{nvidia2025cosmos} as the training-only video denoiser, with eight-step action chunks and eight-frame future targets.

\paragraph{Benchmarks.}
LIBERO comprises four language-conditioned manipulation suites---Spatial, Object, Goal, and Long---with 10 tasks and 500 demonstrations each \citep{liu2023libero}. We pool the training data from all four suites and train a single policy across all 40 tasks. Evaluation uses multiple environment seeds and 50 episodes per task, giving 2{,}000 episodes in total. We report success for each suite together with the unweighted suite mean. For SIMPLER, we jointly train the policy on the BridgeV2~\citep{walke2023bridgedata} and Fractal~\citep{brohan2023rt1} real-robot datasets and evaluate it in a visually matched simulator \citep{li2025simpler}, following the official WidowX protocol on spoon-on-towel, carrot-on-plate, block-stacking, and eggplant-in-basket, using the 24 prescribed object-pose configurations per task and averaging over configurations and seeds; we additionally evaluate the same policy on the standard Google Robot visual-matching tasks.

\paragraph{R1~Pro real-world setup.}
We control the right arm of a Galaxea R1~Pro from a cropped head-mounted fisheye view with an 8-dimensional joint-and-gripper action, the crop removing large background regions while retaining the arm and workspace. The robot places a banana, mango, strawberry, or lemon into a basket. We run 24 trials per task, for 96 evaluation trials per policy, randomizing object--basket positions and the order in which policies are evaluated.

\paragraph{Baselines and protocol.}
Our matched VLA baseline is Qwen-GR00T: the same Qwen3-VL backbone and GR00T-style action head as \method, but with the action expert reading $h_t$ directly and with neither a World Adapter nor a training-time video branch. Table~\ref{tab:libero} and Table~\ref{tab:simpler} further compare against published VLA and WAM results under the official protocols. The $\pi_{0.5}$ variants with and without embodied pretraining \citep{black2025pi05} contrast large-scale pretraining followed by fine-tuning on limited demonstrations against training on those demonstrations alone. Success rate is the primary metric. We measure latency for \method and $\pi_{0.5}$ on a 24~GB RTX~5090~D; the remaining latency values in Table~\ref{tab:libero} are quoted from their source papers and should be interpreted as contextual comparisons.

\subsection{LIBERO}

\method averages 98.2\% across the four LIBERO suites (Table~\ref{tab:libero}). Without embodied pretraining, \method exceeds the 4B StarVLA by 0.4 points with a 2B backbone. World2Act, which likewise discards its world model at deployment, reaches 98.1\% only with embodied pretraining and a 3B backbone. The WAMs with higher averages, DiT4DiT and Cosmos Policy, require 136 and 610~ms per chunk versus 61.85~ms for \method. Measured on identical hardware, \method stays within $1.1\times$ of $\pi_{0.5}$ (61.85 vs.\ 56.32~ms per chunk) while improving the suite average by 1.3 points.

The Long suite is more informative, since its multi-stage tasks give sequential errors room to accumulate. \method attains 97.0\%, within 0.6 points of the highest reported value (97.6\%), despite using neither embodied action pretraining nor an online generative backbone.

\subsection{Ablation Studies}

Table~\ref{tab:ablation} isolates four design factors under the LIBERO protocol.

\begin{table}[htbp]
\centering
\small
\setlength{\tabcolsep}{5.0pt}
\renewcommand{\arraystretch}{1.05}
\resizebox{\textwidth}{!}{%
\begin{tabular}{@{}lccccccccc@{}}
\toprule
& \multicolumn{4}{c}{Design factor} & \multicolumn{5}{c}{LIBERO success rate (\%)} \\
\cmidrule(lr){2-5}\cmidrule(l){6-10}
Variant & Video sup. & Excl. routing & Learned query & Edge anchor & Spatial & Object & Goal & Long & Avg. \\
\midrule
Qwen-GR00T & \xmark & \xmark & \xmark & \xmark & 97.6 & \textbf{98.8} & \textbf{98.2} & 92.8 & 96.9 \\
\addlinespace[1.5pt]
w/o wm & \xmark & \cmark & \cmark & \xmark & 98.6 & 98.4 & 97.5 & 95.0 & 97.4 \\
w/ VLM bypass & \cmark & \xmark & \cmark & \cmark & 98.8 & 98.6 & 97.2 & 94.1 & 97.2 \\
FFN adapter & \cmark & \cmark & \xmark & \cmark & 98.4 & 98.6 & 97.6 & 93.4 & 97.0 \\
\addlinespace[1.5pt]
RGB anchor & \cmark & \cmark & \cmark & \xmark & 98.2 & 98.0 & 97.2 & 91.5 & 96.2 \\
\midrule
\method (ours) & \cmark & \cmark & \cmark & \cmark & \textbf{99.6} & \textbf{98.8} & 97.4 & \textbf{97.0} & \textbf{98.2} \\
\bottomrule
\end{tabular}}
\caption{Controlled LIBERO ablations under the protocol of Table~\ref{tab:libero}. Each factor column marks one design choice: future-video supervision from the jointly fine-tuned video denoiser (\emph{Video sup.}), exclusive action conditioning on the world tokens $q_t$ rather than the full VLM sequence (\emph{Excl.\ routing}), query-based extraction rather than a parameter-matched FFN (\emph{Learned query}), and Canny first-frame conditioning rather than native RGB (\emph{Edge anchor}). Qwen-GR00T is the VLA baseline; each subsequent ablation changes one operative factor from the full method, with the edge anchor inapplicable when video supervision is removed. Bold marks the best value in each column; the \method row reproduces Table~\ref{tab:libero}.}
\label{tab:ablation}
\end{table}

\begin{figure}[htbp]
  \centering
  \includegraphics[width=0.95\textwidth]{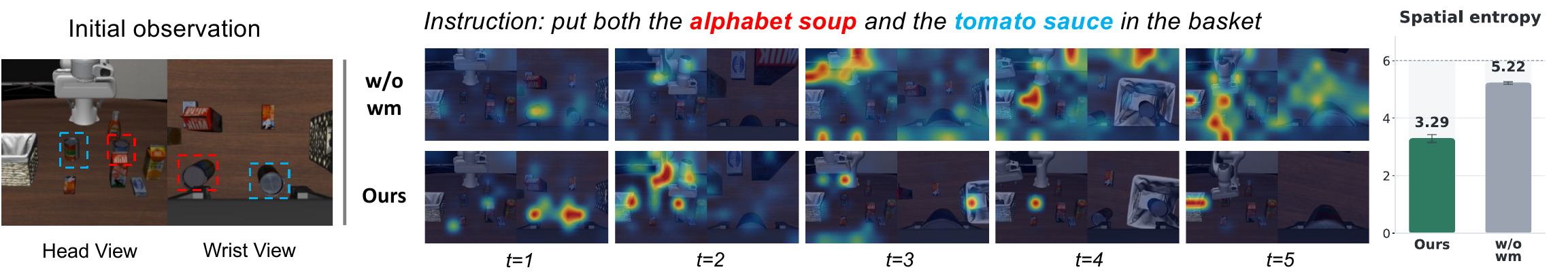}
  \caption{\textbf{Dynamics-aware attention from world tokens.} World Adapter cross-attention from $q_t$ on a LIBERO Long episode (\emph{put both the alphabet soup and the tomato sauce in the basket}), with main-view spatial entropy on the right, averaged over one episode per LIBERO Long task (lower is more focused). Ours concentrates in a task-phase-dependent way; \emph{w/o wm} remains diffuse.}
  \label{fig:attn}
\end{figure}

\paragraph{System-level comparison.}
Relative to Qwen-GR00T, \method improves the average (98.2\% vs.\ 96.9\%), with the gap concentrated on Long (97.0\% vs.\ 92.8\%); the near-saturated suites move less and inconsistently, so we read Long as the informative column.

\paragraph{Mechanistic ablations.}
The next three rows change one factor at a time; all separate only on Long. \emph{w/o wm} removes only the training-time predictive objective; Long falls to 95.0\%, placing the World Adapter alone between Qwen-GR00T and the full method. \emph{w/ VLM bypass} keeps the video branch but restores direct access to $h_t$, falling to 94.1\%---below the variant with no video branch at all. \emph{FFN adapter} replaces query-based extraction with a matched-parameter feed-forward network, reaching 93.4\%. One account is that aligning understanding, prediction, and decision-making calls for a dedicated, sufficiently expressive module learned from scratch, rather than a transformation of the VLM's existing representation structure. Shaped from the start by the action and video objectives, the learned queries grow into a shared interface across the three roles. The FFN extractor, by contrast, merely reshapes VLM features, so the predictive gradient perturbs the representation the action expert relies on.

The ordering matters more than the margins: both variants that keep world-model supervision but break the interface fall below \emph{w/o wm}, so predictive supervision does not help unconditionally. When the action expert can bypass the supervised representation, or extraction lacks learned queries, the video objective competes with action prediction instead of shaping it. Exclusive routing and query-based extraction are thus the mechanism that makes training-time world modeling useful, not implementation details.

\paragraph{Structural-anchor comparison.}
The RGB anchor row changes only the structural anchor: the denoiser receives its native RGB first frame instead of the Canny edge map, while video supervision, exclusive routing through $q_t$, and query-based extraction remain unchanged. Long success falls from 97.0\% to 91.5\%, supporting appearance-suppressed conditioning: a full RGB anchor lets the denoiser rely on appearance cues instead of recovering task-relevant content from the world tokens. In all rows, the video-model components remain training-only.

\paragraph{Attention analysis.}
Figure~\ref{fig:attn} examines how future-video supervision shapes the world tokens on a LIBERO Long episode. We visualize World Adapter cross-attention from $q_t$ onto the head and wrist views. With world-model supervision, the maps are sharper and phase-dependent: at $t{=}1,4$, when the gripper has not yet secured a target object, attention concentrates on the instructed cans, and after a grasp ($t{=}2,3,5$) it shifts to the destination basket; without supervision it remains diffuse and does not track this approach--place structure. To quantify concentration, we report the spatial entropy of the main-view attention over the $8{\times}8$ patch grid, averaged over one episode per LIBERO Long task: \method averages $3.29$\,bits versus $5.22$ for \emph{w/o wm} (uniform is $\log_2 64{=}6$), and is lower at every one of the $32$ paired timesteps on the visualized episode. This contrast indicates that the predictive objective shapes $q_t$---the action expert's sole visual-language context---toward dynamics-relevant structure, consistent with the Long-suite drop under \emph{w/o wm}.

\subsection{SIMPLER}

\begin{table}[htbp]
\centering
\footnotesize
\setlength{\tabcolsep}{2.4pt}
\renewcommand{\arraystretch}{1.0}
\resizebox{\textwidth}{!}{%
\begin{tabular}{@{}lccccccccccc@{}}
\toprule
& & \multicolumn{5}{c}{WidowX Robot (VM)} & \multicolumn{5}{c}{Google Robot (VM)} \\
\cmidrule(lr){3-7}\cmidrule(l){8-12}
Method & Backbone & Spoon & Carrot & Block & Eggpl. & Avg. & Pick & Move & Drawer & Place & Avg. \\
\midrule
$\pi_{0}$ \citep{black2024pi0} & 3B & 29.1 & 0.0 & 16.6 & 62.5 & 27.1 & 72.7 & 65.3 & 38.3 & -- & 58.8 \\
GR00T-N1.5 \citep{bjorck2025gr00t} & 3B & 75.3 & 54.3 & \textbf{57.0} & 61.3 & 61.9 & 51.7 & 54.0 & 27.8 & 7.4 & 35.2 \\
OpenVLA-OFT \citep{kim2025openvlaoft} & 7B & 34.2 & 30.0 & 30.0 & 72.5 & 41.8 & -- & -- & -- & -- & 63.0 \\
SpatialVLA \citep{qu2025spatialvla} & 3B & 16.7 & 25.0 & 29.2 & \textbf{100.0} & 42.7 & 86.0 & 77.9 & 57.4 & -- & 75.1 \\
VLA-JEPA \citep{sun2026vlajepa} & 2B & 75.0 & 70.8 & 12.5 & 70.8 & 57.3 & 88.3 & 64.1 & 59.3 & 49.1 & 65.2 \\
$\pi_{0.5}$ \citep{black2025pi05} & 2B & 49.3 & 64.7 & 44.7 & 69.7 & 57.1 & -- & -- & -- & -- & 72.7 \\
StarVLA \citep{ye2026starvla} & 4B & 79.7 & 59.8 & 22.8 & 98.5 & 65.2 & 90.1 & \textbf{82.6} & 56.3 & 68.7 & 74.3 \\
Qwen-GR00T \citep{ye2026starvla} & 4B & \textbf{83.0} & 59.4 & 18.8 & \textbf{100.0} & 65.3 & -- & -- & -- & -- & 75.3 \\
\midrule
\method (ours) & 2B & 74.0 & \textbf{85.0} & 32.0 & 95.0 & \textbf{71.5} & \textbf{91.0} & 65.0 & \textbf{78.7} & \textbf{93.5} & \textbf{82.1} \\
\bottomrule
\end{tabular}}
\caption{SIMPLER VM success (\%) on WidowX and Google Robot~\citep{ye2026starvla,sun2026vlajepa,qu2025spatialvla}. Google tasks are Pick Coke Can, Move Near, Open/Close Drawer, and Open Top Drawer + Place Apple; reported Google averages may omit Place. -- denotes unavailable; bold marks the best value in each column.}
\label{tab:simpler}
\end{table}

Having isolated the design factors on LIBERO, we next evaluate the resulting policy's system-level competitiveness when trained on real-robot data and evaluated in a visually matched simulator. \method achieves 71.5\% average visual-matching success on SIMPLER-WidowX (Table~\ref{tab:simpler}), the highest Avg.\ among the listed methods; with a 2B VLM backbone, it exceeds the published 4B Qwen-GR00T result by 6.2 points. Because SIMPLER evaluates real-robot-trained policies in simulation, success here depends on features that survive a change of rendering. The advantage is broad-based rather than concentrated in one task: carrot-on-plate tops the table, eggplant-in-basket approaches the best entries, and spoon-on-towel sits mid-table. Block-stacking remains the weakest task, though it is hard for every listed method---none exceeding 57\%---and we make no claim of an advantage on precise stacking.

On the Google Robot embodiment, the same policy attains the highest average (82.1\%), with the best entries on Pick Coke Can, Open/Close Drawer, and Place Apple; Move Near goes to StarVLA. Because several reported Google averages omit the Place task, we treat the per-task columns as the more reliable comparison.

\subsection{Real-World Manipulation}

\begin{figure}[htbp]
  \centering
  \begin{minipage}[c]{0.56\textwidth}
    \includegraphics[width=\linewidth]{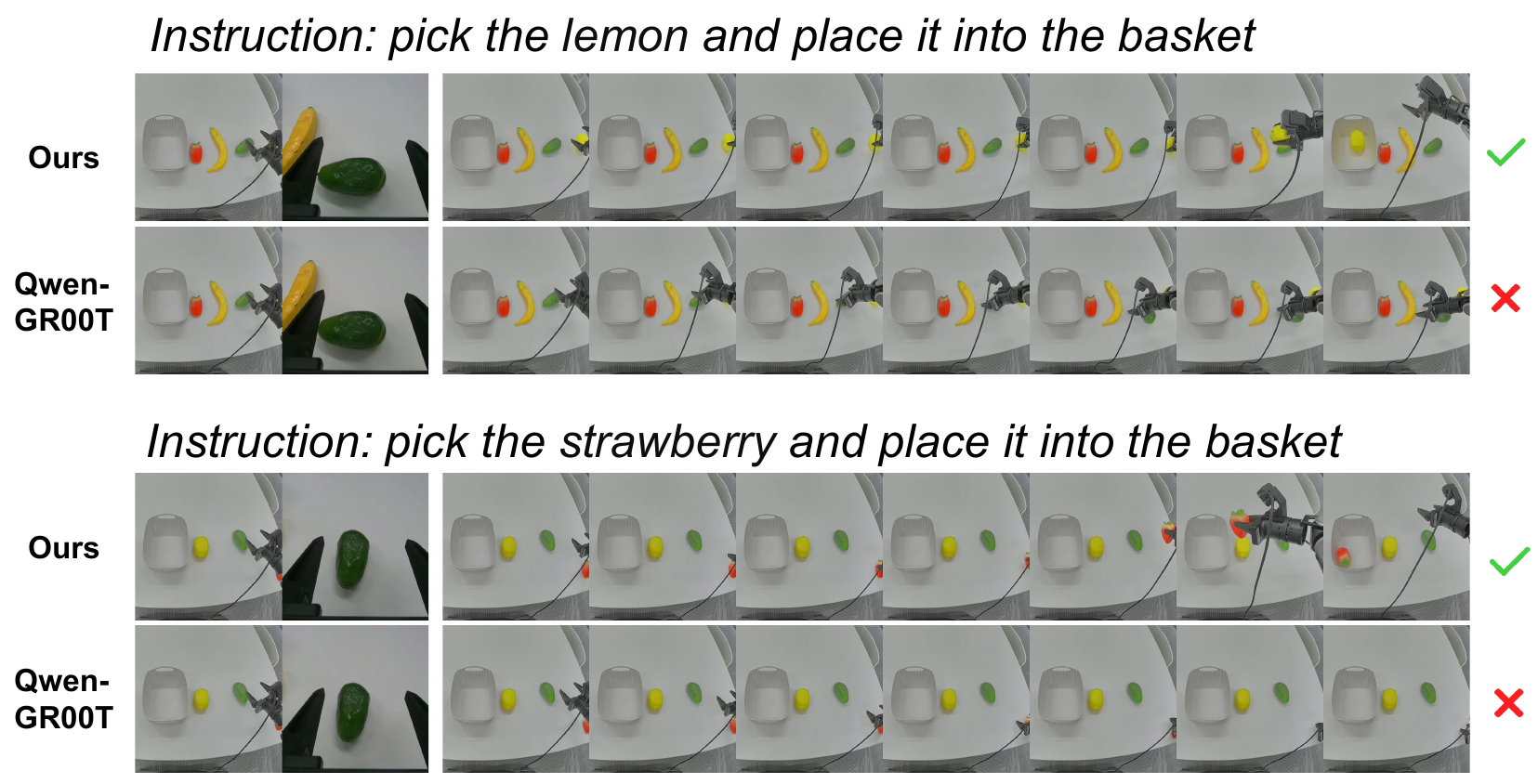}
  \end{minipage}\hfill
  \begin{minipage}[c]{0.40\textwidth}
    \captionsetup{justification=RaggedRight,singlelinecheck=false}%
    \caption{\textbf{Representative R1 Pro rollouts.} For lemon and strawberry, the initial head/wrist views are followed by seven uniformly sampled head-camera frames. \method completes both placements; Qwen-GR00T does not. These examples are illustrative; Table~\ref{tab:real} reports aggregate success.}
  \label{fig:real_qual}
  \end{minipage}
\end{figure}

\begin{table}[htbp]
\centering
\footnotesize
\setlength{\tabcolsep}{2.2pt}
\begin{tabular*}{0.82\textwidth}{@{\extracolsep{\fill}}lccccc@{}}
\toprule
Policy & Lemon & Straw. & Mango & Banana & Avg. \\
\midrule
$\pi_{0.5}$ (w/o pretrain) & 12.5 & 20.8 & 12.5 & 16.7 & 15.6 \\
Qwen-GR00T & 54.2 & 58.3 & 75.0 & 50.0 & 59.4 \\
\method & \textbf{70.8} & \textbf{83.3} & 75.0 & \textbf{75.0} & \textbf{76.0} \\
\midrule
\textcolor{gray}{$\pi_{0.5}$ (pretrained)} & \textcolor{gray}{100.0} & \textcolor{gray}{100.0} & \textcolor{gray}{100.0} & \textcolor{gray}{100.0} & \textcolor{gray}{100.0} \\
\bottomrule
\end{tabular*}
\caption{R1~Pro closed-loop success (\%), 24 trials per object; Avg.\ pools 96 trials. Bold marks the best without embodied pretraining; gray row is context.}
\label{tab:real}
\end{table}

Pooled over the four tasks, \method succeeds on 76.0\% of trials against 59.4\% for Qwen-GR00T (Table~\ref{tab:real}). Strawberry and banana account for most of the difference (25.0 points each), lemon for 16.6, and mango is unchanged, where Qwen-GR00T is already strongest and headroom is smallest, so we rest the claim on the 96 pooled trials.

The two $\pi_{0.5}$ rows are context rather than a matched comparison: trained only on the same demonstrations, $\pi_{0.5}$ reaches 15.6\%, while its embodied-pretrained checkpoint solves every trial, reflecting how embodied pretraining already covers near-in-distribution pick-and-place; the controlled comparison remains Qwen-GR00T.

Figure~\ref{fig:real_qual} shows representative rollouts on lemon and strawberry under randomized scene layouts, where the target fruit is often barely visible initially. Qwen-GR00T fails in two characteristic ways---homing in on a wrong object, or losing the correct fruit to an insecure grasp---whereas \method localizes the instructed fruit from the first frames and completes the placement, consistent with its SIMPLER placement strength.

\section{Conclusion}

We presented \method, an embodied policy architecture that obtains the representational benefits of world modeling without inheriting its inference cost. Its central mechanism is a fixed set of world tokens that jointly condition future-video denoising during training and exclusively condition the action expert; at deployment the video branch is discarded entirely, leaving a policy that runs at VLA-level latency. Across LIBERO, SIMPLER, and a real-robot R1~Pro protocol, this design proves both effective and efficient. Our analysis further clarifies why it works: predictive supervision improves control only when it shapes the representation the policy actually consumes, which is precisely what exclusive routing and query-based extraction enforce, and attention visualizations suggest that the resulting tokens acquire dynamics-relevant structure from the world model. Limitations remain---the world model adds considerable training-time cost, and the Canny anchor is hand-designed---both motivating lighter, learned alternatives.  Future work will explore large-scale joint training of the World Adapter across diverse robot datasets, embodiments, and tasks, with the goal of learning a more general world-token
interface for embodied control.

\clearpage
\appendix
\renewcommand{\thefigure}{A\arabic{figure}}
\renewcommand{\thetable}{A\arabic{table}}
\setcounter{figure}{0}
\setcounter{table}{0}
\setcounter{section}{0}

\section*{Appendix}
\addcontentsline{toc}{section}{Appendix}

\section{Implementation Details}
\label{supp:impl}

\paragraph{Optimization.}
All benchmarks share one optimization recipe. We use AdamW with
$\beta=(0.9,\,0.95)$, $\epsilon=10^{-8}$, and weight decay $10^{-8}$, under a
cosine schedule with linear warmup. Learning rates are assigned per parameter
group according to whether the group is adapted from a pretrained checkpoint or
trained from initialization: $1\times10^{-5}$ for the pretrained
vision-language model, $2\times10^{-5}$ for the jointly fine-tuned video
denoiser, and $1\times10^{-4}$ for the World Adapter, the world-model
projection, and the action expert. The smallest rate on the backbone limits
drift from its pretrained visual-language grounding; the video denoiser is
likewise adapted rather than trained from scratch and takes an intermediate
rate; the from-scratch modules take the largest rate.

\paragraph{Structural anchor.}
The anchor is computed on the primary view of the first frame at the video
branch's own resolution. We convert the frame to grayscale, apply the Canny
detector with hysteresis thresholds $60$ and $140$, blur the resulting binary
edge map with an isotropic Gaussian of $\sigma=2.5$, and replicate the single
channel to RGB before encoding it as the first-frame latent.

\paragraph{Training scale.}
All policies are trained on H100 GPUs with 16 samples per device and no
gradient accumulation. The LIBERO policy is trained for 70K steps at an
effective batch size of 128 across 8 GPUs; the SIMPLER policy, which covers the
larger BridgeV2 and Fractal mixture, for 140K steps at an effective batch size
of 256 across 16 GPUs; and the R1~Pro policy for 30K steps at an effective
batch size of 128 across 8 GPUs. On all three benchmarks the matched
Qwen-GR00T baseline is trained at an identical step count and batch size, so
every comparison against it holds the data and optimization budget fixed and
varies only the architecture. This training
hardware is not the hardware used for the latency measurements in
Section~\ref{supp:latency}.

\paragraph{Deployment footprint.}
The World Adapter uses $K{=}256$ world tokens and $L{=}12$ blocks at width
$d{=}2048$ with 8 attention heads. Each block applies cross-attention to the
vision-language sequence, self-attention over the world tokens, and a
feed-forward layer whose inner dimension equals the width, which is $10d^2$
parameters per block and approximately 0.5B in total. It is the only component
the deployed policy adds over the matched Qwen-GR00T
baseline~\citep{bjorck2025gr00t}; its runtime cost is measured in
Section~\ref{supp:latency}.

\section{Latency Measurement}
\label{supp:latency}

\paragraph{Protocol.}
All latency values correspond to producing one eight-action chunk rather than a
single action. We measure policy inference only---encoding the current
observations and instruction with the vision-language model, recomputing the
world tokens, and integrating the action flow---and exclude image capture,
observation transport, robot communication, environment stepping, action
execution, and policy-server overhead. Every method we time ourselves uses
LIBERO's~\citep{liu2023libero} two $224\!\times\!224$ camera views, batch size one, and an
eight-action horizon, on a single 24~GB RTX~5090~D unless noted otherwise. We
time on-device with CUDA events, discard the warm-up iterations that absorb
graph compilation and allocator growth, and report the steady-state median. We
report the median rather than the mean because the tail is dominated by
occasional allocator and scheduling stalls that would not be experienced as
control latency in a closed loop.

\begin{figure}[htbp]
  \centering
  \includegraphics[width=0.62\textwidth]{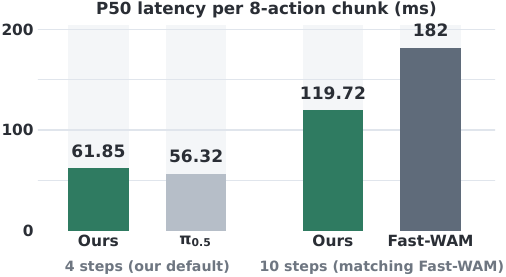}
  \caption{Latency on a single RTX~5090~D with two $224\!\times\!224$ views and
  batch size one. Each pair holds the number of denoising steps and the
  compilation setting fixed, so the two comparisons are like-for-like.}
  \label{supp:fig:latency}
\end{figure}

\paragraph{Measurements.}
We time four policies ourselves---\method, $\pi_{0.5}$~\citep{black2025pi05},
Fast-WAM~\citep{yuan2026fastwam}, and DiT4DiT~\citep{ma2026dit4dit}---by
running the released code and weights of the three baselines under the same
two-view input and the same eight-action chunk. Because we hold the chunk
length at eight rather than adopting each method's native setting, these values
are not intended to reproduce the runtimes reported in the corresponding
papers. Because per-chunk latency also depends on how many denoising steps a
policy takes and on whether its graph is compiled, we group the comparisons by
those two factors rather than placing all methods on one axis
(Figure~\ref{supp:fig:latency}).

Four flow-integration steps is our default and the setting behind every result
in the paper; with a compiled graph and cached conditioning, \method produces
one chunk in 61.85~ms, an action throughput of roughly 129 actions per second.
$\pi_{0.5}$, timed on the same GPU under the same four-step compiled setting
with its own default JAX/XLA compilation and KV cache, produces a chunk in
56.32~ms, placing \method within $1.1\times$ of it. DiT4DiT also denoises in
four steps by default and takes 136~ms, measured on an H100 rather than on our
GPU; if anything this cross-hardware comparison favors DiT4DiT, so we read it
as evidence that \method is not the slower of the two rather than as a
measured speedup.

Fast-WAM is the exception: it denoises in ten steps without compilation. We
therefore re-timed \method in that same ten-step uncompiled configuration
purely to make the comparison fair, and report 119.72~ms against Fast-WAM's
182~ms, a factor of $1.5$; the ten-step number is a like-for-like control, not
a configuration we deploy. Fast-WAM is also the most informative comparison of
the three, being the design closest to ours in intent---video supervision
without test-time future imagination---so reading this gap against our
four-step default instead would overstate it, since part of that difference is
a choice of integration budget rather than a property of the architecture.

\paragraph{World Adapter overhead.}
A comparison against the matched Qwen-GR00T baseline, which shares the same
vision-language model and action expert but computes no adapter, attributes
roughly 10~ms of the 61.85~ms to the World Adapter. The vision-language forward
pass and the four-step flow integration are identical between the two policies,
so this is the adapter's own contribution rather than misattributed shared
cost: at 0.5B parameters the adapter is not the deployment bottleneck.

\paragraph{Comparability.}
The $\pi_{0.5}$, Fast-WAM, and DiT4DiT numbers above are timed under one
protocol; the remaining latency entries in the main text are quoted from
their sources and are not controlled: OpenVLA's 240~ms~\citep{kim2024openvla}
is an A100 measurement of one autoregressively decoded action rather than an
eight-action chunk; OpenVLA-OFT's 112~ms~\citep{kim2025openvlaoft} is an A100
measurement of an eight-step chunk produced by parallel decoding with
continuous regression actions, including a wrist image and proprioceptive
state; $\pi_0$'s 73~ms~\citep{black2024pi0} is total on-board inference on a
consumer RTX~4090, excluding the network latency its off-board configuration
reports separately; and Cosmos Policy's 610~ms~\citep{kim2026cosmospolicy} is
measured on one H100 with five denoising steps, covering action, future-state,
and value predictions generated in parallel over a longer chunk than ours. Any
specific speedup factor against these four should be read as hardware- and
recipe-specific.

\section{World-Model Rollout}
\label{supp:wm}

\begin{figure}[htbp]
  \centering
  \includegraphics[width=\columnwidth]{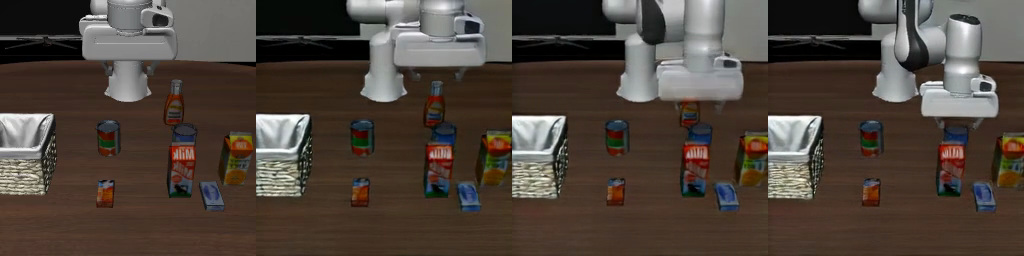}\\[1.2pt]
  \includegraphics[width=\columnwidth]{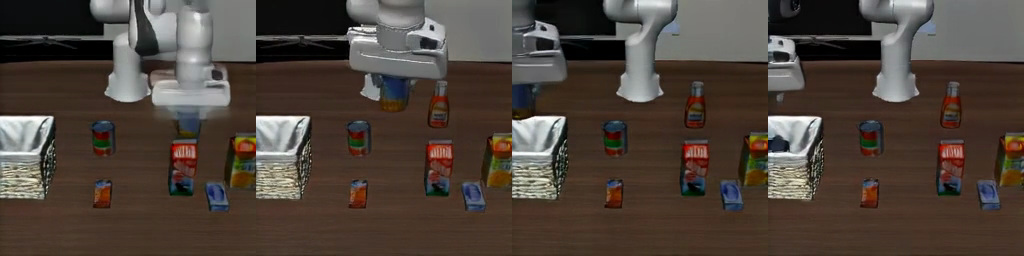}\\[1.2pt]
  \includegraphics[width=\columnwidth]{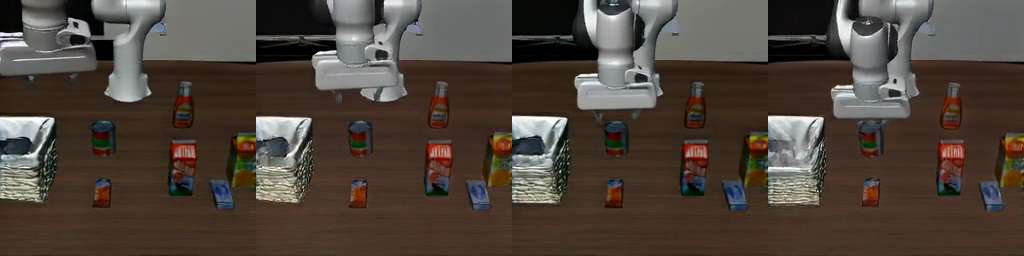}\\[1.2pt]
  \includegraphics[width=\columnwidth]{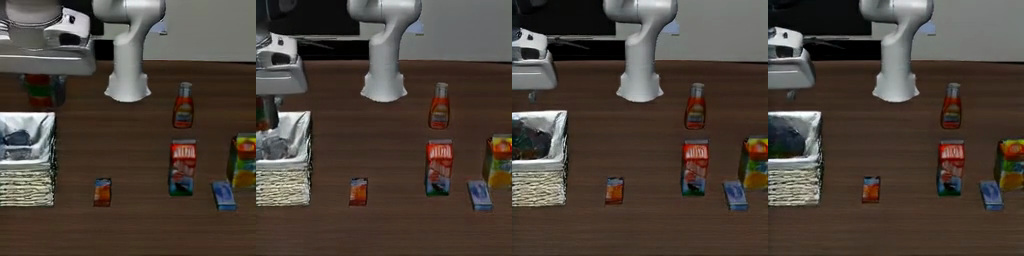}
  \caption{\textbf{Autoregressive world-model rollout.} Sixteen frames sampled
  uniformly from a video rolled out autoregressively by the world-model branch
  from a single observed frame, on the same LIBERO Long episode as Figure~\ref{fig:attn}. Frames read left to right, top row first.}
  \label{supp:fig:wm}
\end{figure}

Figure~\ref{supp:fig:wm} shows the world-model branch rolled out
autoregressively from a single observed frame, on the same LIBERO Long episode
(\emph{put both the alphabet soup and the tomato sauce in the basket}) used for
the attention maps in Figure~\ref{fig:attn} and in
Section~\ref{supp:attn}. Frames are sampled uniformly over the rollout. The
scene layout and the arm stay coherent across the rollout, and the arm moves
through repeated reaching and retracting motions between the objects and the
basket, consistent with the branch having fit the scene dynamics of the task
rather than its static appearance alone. This branch is used only during
training and is discarded at deployment.

\section{Attention Analysis}
\label{supp:attn}

\begin{figure}[htbp]
  \centering
  \includegraphics[width=0.72\textwidth]{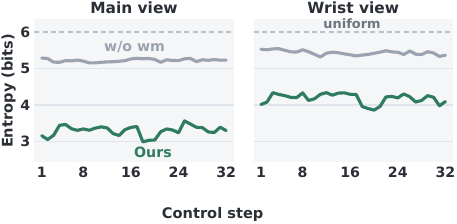}
  \caption{Spatial entropy of World Adapter attention at each control step of
  the LIBERO Long episode visualized in the main text. The dashed line is the
  uniform-attention bound $\log_2 64 = 6$~bits. The separation holds at every
  step on both views rather than at a few frames.}
  \label{supp:fig:entropy}
\end{figure}

\paragraph{Procedure.}
For a given control step we take the cross-attention weights of the World
Adapter's final block, from all 256 world tokens to the vision-language
sequence, average them over the 8 attention heads and over the world tokens,
and restrict the result to the image-token positions of one camera view. Each
view contributes an $8\!\times\!8$ grid of 64 patch tokens, so this yields a
distribution over 64 spatial positions per view and per control step, whose
Shannon entropy is at most $\log_2 64 = 6$~bits. A control step is one policy
invocation: the World Adapter recomputes the world tokens once per action
chunk, so the 32 control steps of the episode in
Figure~\ref{supp:fig:entropy} span 256 environment steps.
We compute this quantity at every control step of one episode per LIBERO Long
task and average across tasks. \method and the \emph{w/o wm} ablation are
evaluated on the same episodes under the same environment seeds, so the two
trajectories are step-aligned and can be compared step by step. The maps
visualized in the main text are these distributions rendered over the
observation and upsampled for display; no smoothing, contrast adjustment, or
per-map renormalization is applied beyond a shared color scale.

\paragraph{Additional metrics.}
The main text reports the main-view entropy contrast as an average. Resolving
it per control step on the visualized episode
(Figure~\ref{supp:fig:entropy}) shows that the separation is not an artifact of
averaging: \method attends more sharply at all 32 control steps on both views
and the two curves never approach each other. An exact paired sign test over
those steps returns its floor value $2^{-32}$ on each view; we report it only
as a description of how uniform the ordering is, since consecutive control
steps of one episode are not independent samples.
The effect is also not confined to the primary
camera---on the wrist view, entropy averages 4.18~bits for \method against 5.44
for \emph{w/o wm}, a smaller but same-signed gap, consistent with the wrist
view carrying less task-discriminative layout information. Nor is it only
visible in the entropy summary: on the main view the single most-attended patch
receives 0.42 of the total attention mass under \method versus 0.10 without
world-model supervision, and the top four of the 64 patches receive 0.63 versus
0.29. All of these are descriptive statistics of where the adapter attends; the
causal claim rests on the ablations in the main text.

\section{Additional Real-World Rollouts}
\label{supp:rollouts}

\newlength{\rlabw}\setlength{\rlabw}{0.070\textwidth}
\newlength{\robsw}\setlength{\robsw}{0.195\textwidth}
\newlength{\rpredw}\setlength{\rpredw}{0.6825\textwidth}
\newlength{\rmarkw}\setlength{\rmarkw}{0.022\textwidth}
\newcommand{\rlab}[1]{\footnotesize\bfseries\shortstack[l]{#1}}

\begin{figure}[htbp]
  \centering
  \setlength{\tabcolsep}{0pt}
  \begin{tabular}{@{}m{\rlabw}@{\hspace{2pt}}m{\robsw}@{\hspace{6pt}}m{\rpredw}@{\hspace{5pt}}m{\rmarkw}@{}}
    \multicolumn{4}{c}{\itshape Instruction: pick the lemon and place it into the basket}\\[2pt]
    \rlab{Ours}            & \includegraphics[width=\robsw]{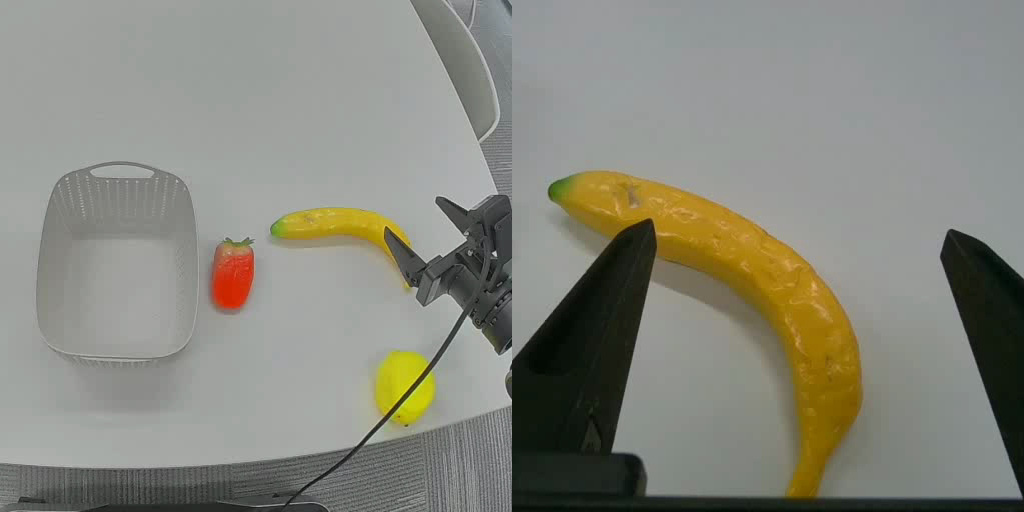}  & \includegraphics[width=\rpredw]{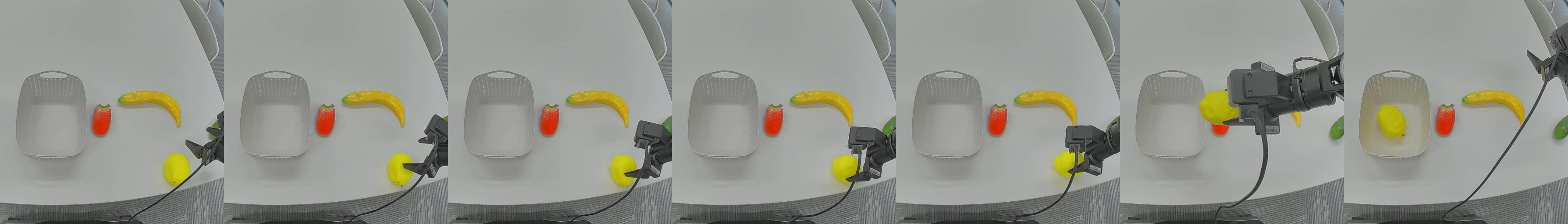}  & \okmark   \\[2pt]
    \rlab{Qwen-\\GR00T}    & \includegraphics[width=\robsw]{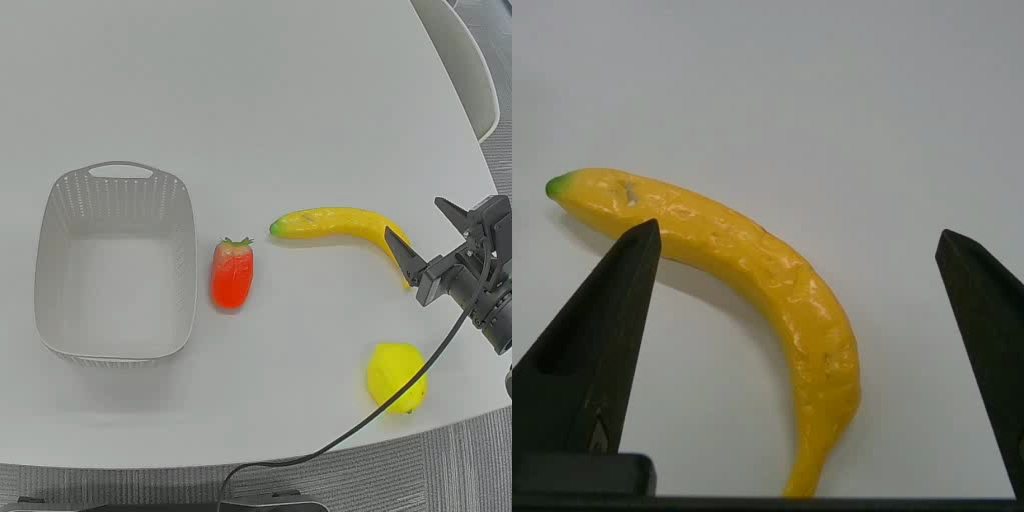}  & \includegraphics[width=\rpredw]{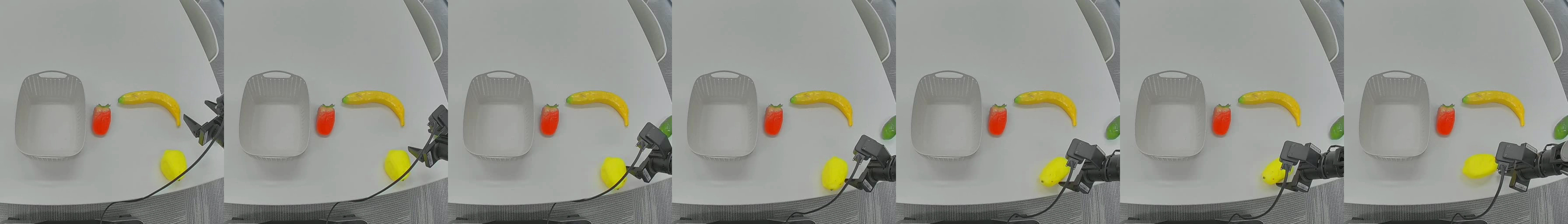}  & \failmark \\[9pt]
    \multicolumn{4}{c}{\itshape Instruction: pick the banana and place it into the basket}\\[2pt]
    \rlab{Ours}            & \includegraphics[width=\robsw]{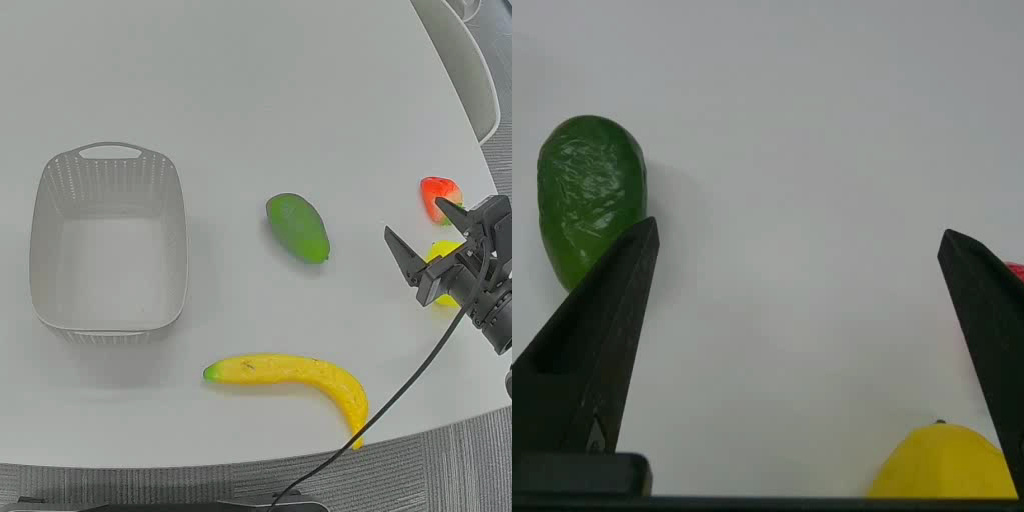} & \includegraphics[width=\rpredw]{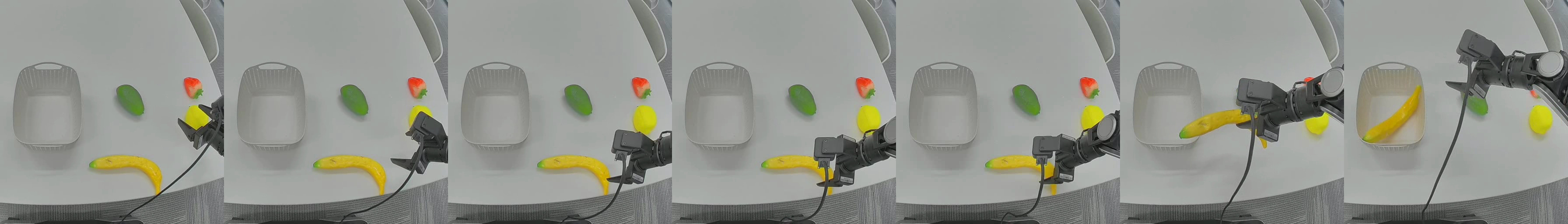} & \okmark   \\[2pt]
    \rlab{Qwen-\\GR00T}    & \includegraphics[width=\robsw]{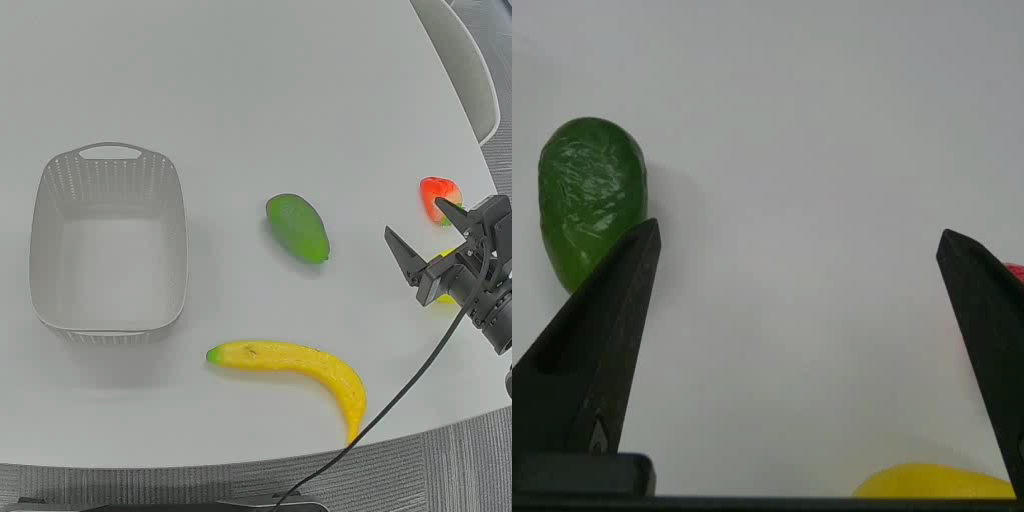} & \includegraphics[width=\rpredw]{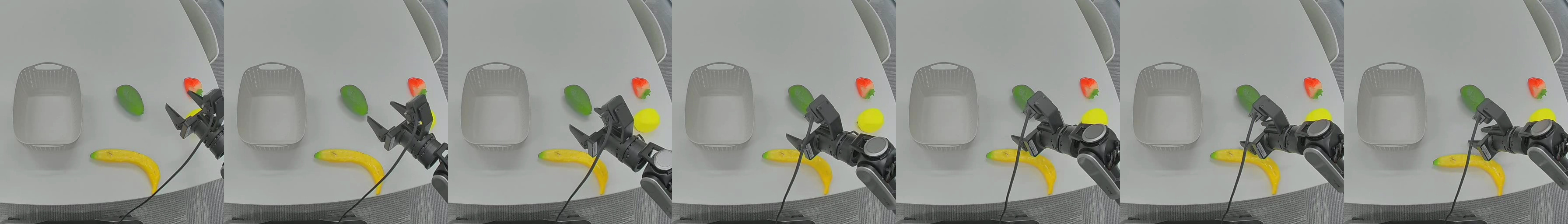} & \failmark \\[9pt]
    \multicolumn{4}{c}{\itshape Instruction: pick the mango and place it into the basket}\\[2pt]
    \rlab{Ours}            & \includegraphics[width=\robsw]{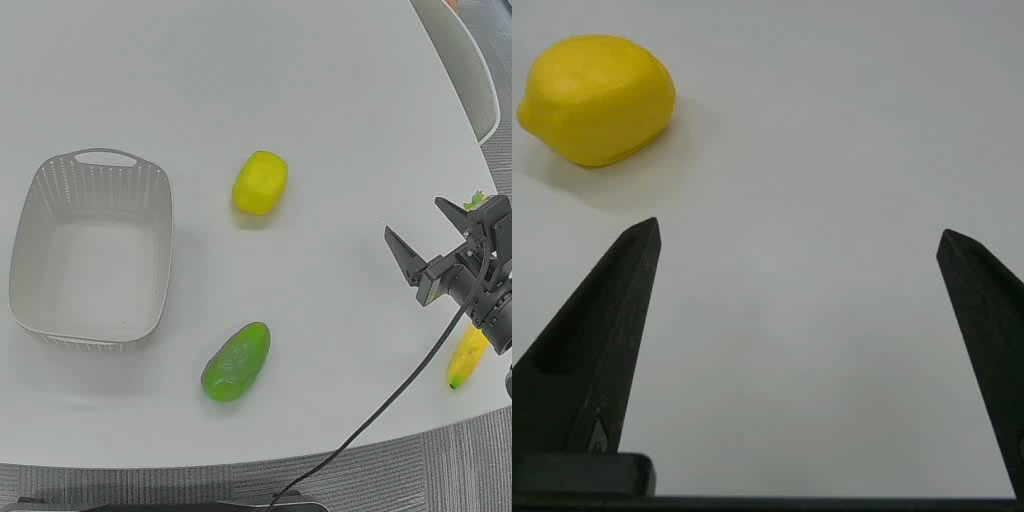}  & \includegraphics[width=\rpredw]{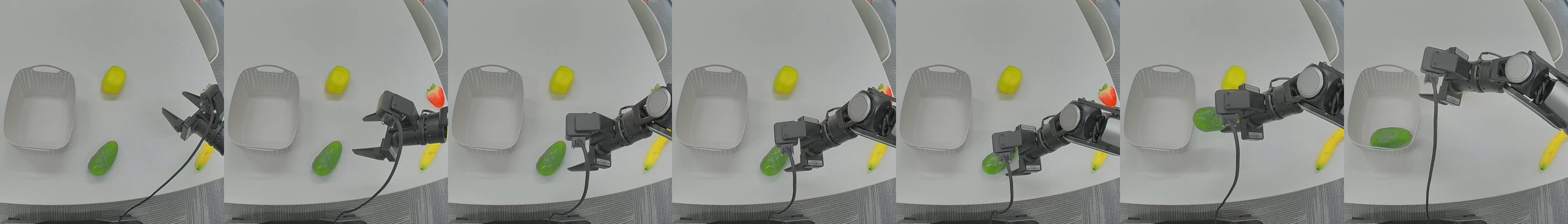}  & \okmark   \\[2pt]
    \rlab{Qwen-\\GR00T}    & \includegraphics[width=\robsw]{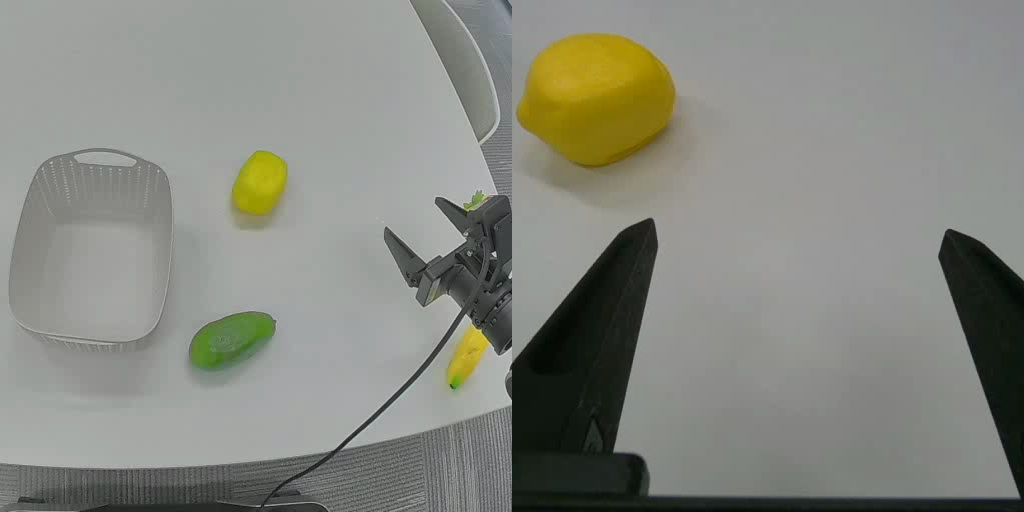}  & \includegraphics[width=\rpredw]{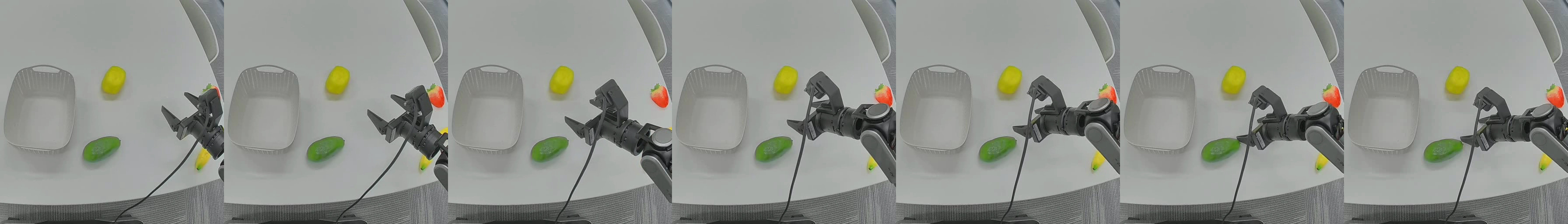}  & \failmark \\
  \end{tabular}
  \caption{\textbf{Additional R1~Pro rollouts.} Further examples on the three
  tasks not shown in the main text. As in the figure in the main text, the two
  leftmost cells are the initial head and wrist views, followed by seven
  head-camera frames sampled uniformly over the rest of the episode. A check
  marks an episode ending with the named object in the basket and a cross one
  that does not. Aggregate success is reported in Table~\ref{tab:real}.}
  \label{supp:fig:rollouts}
\end{figure}

Figure~\ref{supp:fig:rollouts} shows additional R1~Pro rollouts on lemon,
banana, and mango, complementing the lemon and strawberry examples in the main text. Within each task, \method and Qwen-GR00T start from the same object
arrangement, and all six episodes come from the same evaluation session as the
examples in the main text.

\end{document}